\documentclass[conference]{IEEEtran}
\IEEEoverridecommandlockouts

\usepackage[noend]{algpseudocode}
\usepackage[font={small,it}]{caption}
\usepackage{float}
\usepackage{graphicx}
\usepackage{epsfig}
\usepackage{amsmath}
\usepackage{amssymb} 
\usepackage{longtable}
\usepackage{algorithm}
\usepackage{rotating}
\usepackage{multirow}
\usepackage{array}
\usepackage{graphicx}
\usepackage{mathrsfs}
\usepackage{verbatim}
\usepackage{color}
\usepackage{epstopdf}
\usepackage{url}
\usepackage{graphicx}
\usepackage{mathrsfs}
\usepackage{verbatim}
\usepackage{color}
\usepackage[noadjust]{cite}
\usepackage{epstopdf}
\usepackage{url}
\usepackage{subfig}

\usepackage{graphicx}
\usepackage{epsfig}
\usepackage{amsmath}
\usepackage{amssymb} 
\usepackage{longtable}
\usepackage{rotating}
\usepackage{multirow}
\usepackage{array}
\usepackage{graphicx}
\usepackage{mathrsfs}
\usepackage{verbatim}
\usepackage{color}
\usepackage{epstopdf}
\usepackage{url}
\usepackage{graphicx}
\usepackage{mathrsfs}
\usepackage{verbatim}
\usepackage{soul}
\usepackage{epstopdf}
\usepackage{url}

\usepackage{mwe}

\usepackage{pgfplots,tikz-3dplot}
\usepackage{amsmath} % assumes amsmath package installed
\usepackage{amssymb}  % assumes amsmath package installed
\usepackage{bbm}
\usepackage{color}
\usepackage{graphics}

\usepackage{epsfig}
\usepackage{epstopdf}
\usepackage{amsfonts}
\usepackage{mathtools}
\usepackage{bm}
\usepackage{fancyhdr}
\usepackage{framed}
\usepackage{float}
\usepackage{verbatim}
\usepackage{tikz}
\usepackage{relsize}
\usepackage{mathrsfs}
\usepackage{todonotes}
\usepackage{pgfplots}
\usepackage{grffile}
\usepackage{color}
\usetikzlibrary{arrows}

\usepackage{todonotes}
\usepackage[font={footnotesize},labelfont=bf]{caption}

\usepackage{setspace}

\renewcommand{\baselinestretch}{0.92}

\newtheorem{th1}{\bf{Theorem}}

\newcommand{\bs}{\boldsymbol}

\newtheorem{assumption}{\bf{Assumption}}
\newtheorem{Property}{\textbf{Property}}

\newcolumntype{C}[1]{>{\centering\let\newline\\\arraybackslash}m{#1}}
\newcommand{\myfrac}[3][0pt]{\genfrac{}{}{}{}{\raisebox{#1}{$#2$}}{\raisebox{-#1}{$#3$}}}

\title{\Large \bf Decentralized Impedance Control for Cooperative Manipulation of Multiple Underwater Vehicle Manipulator Systems under Lean Communication}

\author{Shahab Heshmati-Alamdari, Charalampos P. Bechlioulis, George C. Karras and Kostas J. Kyriakopoulos\vspace{-8mm}% <-this % stops a space
\thanks{The authors are with the Control Systems Lab, Department of Mechanical Engineering, National Technical University of Athens, 9 Heroon Polytechniou Street, Zografou 15780, Greece. Emails: {\tt\small \{shahab,chmpechl,karrasg,kkyria@mail.ntua.gr\}}. }%
}

\begin{document}
\maketitle \thispagestyle{empty} \pagestyle{empty}

\setlength{\belowdisplayskip}{4pt}
\begin{abstract} 
This paper addresses the problem of cooperative object transportation for multiple Underwater Vehicle Manipulator Systems (UVMSs) in a constrained workspace with static obstacles, where the coordination relies solely on implicit communication arising from the physical interaction of the robots with the commonly grasped object. We propose a novel distributed leader-follower architecture, where the leading UVMS, which has knowledge of the object's desired trajectory, tries to achieve the desired tracking behavior via an impedance control law, navigating in this way, the overall formation towards the goal configuration while avoiding collisions with the obstacles. On the other hand, the following UVMSs estimate the object's desired trajectory via a novel prescribed performance estimation law and implement a similar impedance control law. The feedback relies on each UVMS's force/torque measurements and no explicit data is exchanged online among the robots. Moreover, the control scheme adopts load sharing among the UVMSs according to their specific payload capabilities. Finally, various simulation studies clarify the proposed method and verify its efficiency. \vspace{2mm}

\end{abstract}

\section{Introduction}
During the last decades, Unmanned Underwater Vehicles (UUVs) have been widely used in various applications such as marine science (e.g., biology, oceanography, archeology) and offshore industry (e.g., ship maintenance, inspection of oil/gas facilities) \cite{Fossen2,heshmati2018robust,heshmati2015robust}. In particular, a vast number of the aforementioned applications, demand the underwater vehicle to be enhanced with intervention capabilities as well \cite{HESHMATIALAMDARI2018}, thus raising increasing interest on Underwater Vehicle Manipulator System (UVMS). For instance, some recent European projects: TRIDENT \cite{Simetti2014364,Ribas20152583}, PANDORA \cite{hurtos2014sonar,heshmati2014self}, and the most recent one DexROV \cite{Gancet2015218}, have boosted significantly the autonomous underwater interaction tasks.

\begin{figure}
	\centering
	\setlength{\fboxsep}{0pt}%
	\setlength{\fboxrule}{1pt}%
	{\includegraphics[width=0.28\textwidth]{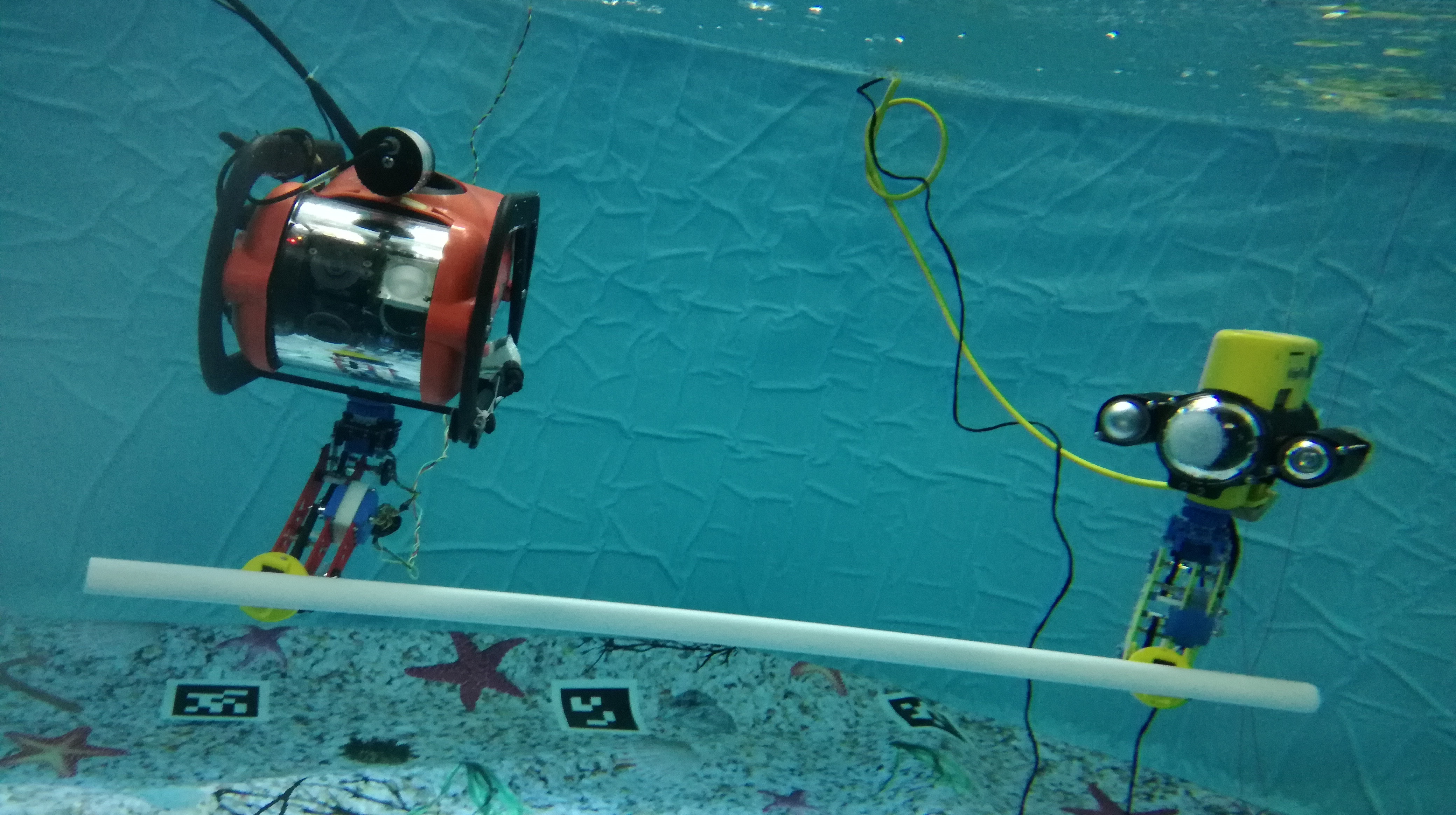}}
	\caption{Two custom-made UVMSs under cooperative transportation.}
	\label{fig:uvms_frames}
\end{figure}

Most of the underwater manipulation tasks can be carried out more efficiently, if multiple UVMSs are cooperatively involved (see Fig \ref{fig:uvms_frames}). In general, underwater tasks are very demanding, with the most significant challenge being imposed by the strict communication constraints \cite{Marani200915}. In general, the communication of multi-robot systems can be classified in two major categories, namely explicit (e.g., conveying information such as sensory data directly to other robots) and implicit (e.g., the interaction forces between the object and the robot). Even though the inter-robot communication is of utmost importance during cooperative manipulation tasks, employing explicit communication in underwater environment may result in severe performance problems owing to the limited bandwidth and update rate of underwater acoustic devices. Moreover, as the number of cooperating robots increases, communication protocols require complex design to deal with the crowed bandwidth \cite{Stilwell20002358}.  
Cooperative manipulation has been well-studied in the literature, especially the centralized schemes \cite{Nikou2017707}. Despite its efficiency, centralized control is less robust, since all units rely on a central system. On the other hand, the decentralized cooperative manipulation schemes depend usually on either explicit communication interchange among the robots (e.g., online transmission of the desired trajectory \cite{alex_chris_ECC_2018} or off--line knowledge of the objects' trajectory \cite{Dickson19973589}). This demands an accurate common global localization system for all participating robots \cite{Conti2015261}, which is difficult to be achieved in underwater environment. Therefore, the design of decentralized cooperative manipulation algorithms for underwater tasks employing implicit and lean explicit communication becomes apparent. In recent studies \cite{Furferi20161}, potential fields methods were employed  to manage the guidance of UVMSs and the manipulation tasks.  Compelling results towards the same direction have been given in \cite{Simetti2016,Manerikar2015}. In particular, a three-fold decentralized cooperative control strategy is proposed where initially, each robot individually finds out an optimal task space control velocity, which is transfered afterwards among the robots in order to obtain a commonly agreed velocity via a fusion policy. However, employing the aforementioned strategies, requires each robot to communicate with the whole robot team, which consequently restricts the number of robots involved in the cooperative manipulation task owing to bandwidth limitations.
 
In this work, the problem of decentralized cooperative object transportation considering multiple UVMSs in a constrained workspace with static obstacles is addressed. The challenge lays in replacing explicit communication with implicit, by incorporating feedback that results from the physical interaction of the robots with the commonly grasped object (i.e., we assume that each UVMS is equipped with a force/torque sensor attached on its end-effector). The proposed scheme is based on a leader-follower architecture, where the leader, which  is aware of the object's desired trajectory, implements it via an  impedance control law. On the other hand, the followers estimate the desired trajectory in a distributed way, via observing the object motion, and impose a similar impedance law. All impedance laws linearize the dynamics and incorporate coefficients for load sharing. The estimation process is based on the prescribed performance methodology \cite{Bechlioulis20101220} that drives the estimation error to an arbitrarily small residual set. Moreover, we design adaptive control laws in order to compensate for the parametric uncertainty of the UVMSs dynamics as well as the external disturbances. Finally, the proposed scheme exploits information, i.e., force/torque at the end-effector, position and velocity measurements, acquired solely by onboard sensors, avoiding thus any tedious inter-robot explicit communication.

\section{Problem Formulation}
Consider $N+1$ UVMSs rigidly grasping an object in a constrained workspace with static obstacles. We also assume that each UVMS is fully-actuated at its end-effector frame.  It should also be noted that in the proposed scheme, only the leading robot is aware of the obstacles' position in the workspace and the object's desired configuration $\bs{x}^d_o$. On the other hand, the followers estimate locally in a distributed way the object's desired trajectory profile and manipulate the object in coordination with the leader based solely on their own  sensory information. Moreover, we assume that the UVMSs can measure their position and velocity (e.g., employing a  fusion technique based on measurements by various onboard sensors), as well as the interaction forces/torques with the object via a force/toque sensor. Additionally, the geometric parameters of both the UVMSs and the commonly grasped object are considered known, whereas their dynamic parameters are completely unknown. 
\subsection{Kinematics}  
We denote the coordinates of the commonly agreed body-fixed frame on the object as well as the leader's and followers' task space (i.e., end-effector) coordinates by $\bs{x}_O=[\bs{\eta}_{1,O}^\top,\bs{\eta}_{2,O}^\top]^\top$, $\bs{x}_L=[\bs{\eta}_{1,L}^\top,\bs{\eta}_{2,L}^\top]^\top$ and $\bs{x}_{F_i}=[\bs{\eta}_{1,F_i}^\top,\bs{\eta}_{2,F_i}^\top]^\top,~i\in\mathcal{N}=\{1,\ldots,N\}$ respectively. More specifically, $\bs{\eta}_{1,i}=[x_i,y_i,z_i]^\top$ and $\bs{\eta}_{2,i}=[\phi_i,\theta_i,\psi_i]^\top,~ i\in \{O,L,F_1,\ldots,F_N\}$ denote the position and the orientation expressed in Euler angles representation with respect to the inertial frame.  Alternatively, the orientation coordinates $\bs{\eta}_{2,i},~ i\in \{O,L,F_1,\ldots,F_N\}$ expressed in Euler angles may be described by a rotation matrix $\bs{R}_i=[\bs{n}_{i},\bs{o}_{i},\bs{\alpha}_{i}]\in\mathbb{R}^3,~ i\in \{O,L,F_1,\ldots,F_N\}$ \cite{sciavicco2012modelling}:\vspace{0mm}
\begingroup\makeatletter\def\f@size{9.0}\check@mathfonts
\def\maketag@@@#1{\hbox{\m@th\large\normalfont#1}}
\begin{align}
\bs{R}_i\!=\!\begin{bmatrix}
c_{\psi_i} c_{\theta_i}&  c_{\psi_i} s_{\theta_i} s_{\phi_i}\!-\!s_{\psi_i} c_{\phi_i}&  c_{\psi_i} s_{\theta_i} c_{\phi_i}\!+\!s_{\psi_i} s_{\phi_i}\\
s_{\psi_i} c_{\theta_i}&  s_{\psi_i} s_{\theta_i} s_{\phi_i}\!+\!c_{\psi_i} c_{\phi_i}& s_{\psi_i} s_{\theta_i} c_{\phi_i}\!-\!c_{\psi_i} s_{\phi_i}\\
-s_{\theta_i}&   c_{\theta_i} s_{\phi_i} &c_{\theta_i} c_{\phi_i}
\end{bmatrix} \label{rotation}
\end{align}\endgroup 
where $s_\star=\sin(\star)$ and $c_\star=\cos(\star)$.  Let $\bs{q}_i=[\bs{q}^\top_{v,i},~\bs{q}^\top_{m,i}]^\top\in \mathbb{R}^{6+n}, i\in\mathcal{K}=\{L,F_1,\ldots,F_N\}$  be the joint state variables of each UVMS,  where $\bs{q}_{v,i} \in \mathbb{R}^6$ is the vector that involves the position and the orientation of the vehicle and $\bs{q}_{m,i}\in\mathbb{R}^n$ is the vector of the angular positions of the manipulator's joints. 
%  Thus, we have \cite{Fossen2,antonelli}:
%  \begin{equation}
%  \dot{\bs{q}}_{v,i}= \bs{J}^v_i(\bs{q}_{v,i})\bs{v}_i,~ i\in\mathcal{K}\label{eq1}
%  \end{equation}
%  where $\bs{v}_i$ is the velocity of the vehicle expressed in the body-fixed frame and $\bs{J}^v_i(\bs{q}_{v,i})$ is the Jacobian matrix transforming the velocities from the body-fixed to the inertial frame.   
Let also define the object as well as the leader's and followers' end effector generalized velocities by ${\bs{v}}_O=[\dot{\bs{\eta}}_{1,O}^\top,\bs{\omega}_O^\top]^\top$, ${\bs{v}}_L=[\dot{\bs{\eta}}_{1,L}^\top,\bs{\omega}_L^\top]^\top$  and ${\bs{v}}_i=[\dot{\bs{\eta}}_{1,i}^\top,\bs{\omega}_i^\top]^\top,~ i\in \{F_1,\ldots,F_N\}$, where $\dot{\bs{\eta}}_{1,i}$ and $\bs{\omega}_i$ denote the linear and angular velocity respectively. Without any loss of generality, for the augmented UVMS system we get \cite{antonelli}:\vspace{-5mm}
\begingroup\makeatletter\def\f@size{9.5}\check@mathfonts
\def\maketag@@@#1{\hbox{\m@th\large\normalfont#1}}
\begin{equation}
~~~~~~~~~~~~~~~~~~~~~~{\bs{v}}_i= \bs{J}_i({\bs{q}}_i)\boldsymbol{\zeta}_i,~ i\in\mathcal{K}\label{eq222}
\end{equation}\endgroup
where $\boldsymbol{\zeta}_i=[\bs{\upsilon}_i^\top,\dot{\bs{q}}_{m,i}^\top]^\top \in \mathbb{R}^{6+n}$  is the velocity vector involving the body velocities of the vehicle as well as the joint velocities of the manipulator with $\bs{\upsilon}_i$ denoting the velocity of the vehicle expressed in the body-fixed frame and $ \bs{J}_i(\bs{q}_i)$ denoting the geometric Jacobian matrix \cite{antonelli}. Furthermore, owing to the rigid grasp of the object and since the object geometric parameters are considered known,
each UVMS can compute the object's position w.r.t the inertial frame $\{I\}$. Furthermore, along with the fact that $\bs{\omega}_i=\bs{\omega}_O,~i\in \mathcal{K}$, one obtains:\vspace{-5mm}
\begingroup\makeatletter\def\f@size{9.5}\check@mathfonts
\def\maketag@@@#1{\hbox{\m@th\large\normalfont#1}}
\begin{align}
{\bs{v}}_i= \bs{J}_{iO}{\bs{v}}_O,~ i\in\mathcal{K}\label{eq4}
\vspace{-2mm}\end{align}\endgroup 
where $\bs{J}_{iO},~i \in\mathcal{K}$ denotes the Jacobian from the end-effector of each UVMS to the object's center of mass, that is defined as:\vspace{-1mm}
\begingroup\makeatletter\def\f@size{9.5}\check@mathfonts
\def\maketag@@@#1{\hbox{\m@th\large\normalfont#1}}
\begin{align*}
\bs{J}_{iO}=\arraycolsep=5.4pt \left[
\begin{array}{ccc}
\bs{I}_{3\times 3} & -\bs{S}(\bs{l}_i)\\
\bs{0}_{3\times 3} & \bs{I}_{3\times 3}
\end{array} \right]\in \mathbb{R}^{6\times 6},~~i \in\mathcal{K}%\label{J_oi}
\end{align*}\endgroup
where $\bs{S}(\bs{l}_i)$ is the skew-symmetric matrix of the constant relative position of the end-effector w.r.t the object $\bs{l}_i=[l_{ix},l_{iy},l_{iz}]^\top$. Notice that $\bs{J}_{iO},~i\in\mathcal{K}$ are always full-rank owing to the grasp rigidity and hence obtain a well defined inverse. Thus, the object's velocity can be easily computed via the inverse of \eqref{eq4}. Moreover, from \eqref{eq4}, one obtains:\vspace{-2mm}
\begingroup\makeatletter\def\f@size{9.5}\check@mathfonts
\def\maketag@@@#1{\hbox{\m@th\large\normalfont#1}}
\begin{align}
\dot{\bs{v}}_i=\bs{J}_{iO}\dot{\bs{v}}_O+\dot{\bs{J}}_{iO}\bs{v}_O,~i\in\mathcal{K}\label{dotv_obj}\end{align}\endgroup

\vspace{-5mm}
 \subsection{Dynamics}\vspace{-1mm}
\textbf{UVMS Dynamics:} The dynamics of a UVMS after straightforward algebraic manipulations can be written as \cite{antonelli}: \vspace{-1mm}
\begingroup\makeatletter\def\f@size{9.5}\check@mathfonts
\def\maketag@@@#1{\hbox{\m@th\large\normalfont#1}}
\begin{align}
\!\bs{{M}}_{q_i}(\bs{q}_i)\dot{\bs{\zeta}}_i\!+\!\bs{{C}}_{q_i}({\bs{\zeta}}_i,\bs{q}_i)&{\bs{\zeta}}_i\!+\!\bs{{D}}_{q_i}({\bs{\zeta}}_i,\bs{q}_i){\bs{\zeta}}_i\!+\!\bs{{g}}_{q_i}(\bs{q}_i)\!\nonumber\\
&+\!\bs{d}_{q_i}(\bs{\zeta}_i,\bs{q}_i,t)=\!\boldsymbol{\tau}_i\!+\!{\bs{J}_i}^\top\boldsymbol{\lambda}_i\label{eq6}
\end{align}\endgroup
for $i\in\mathcal{K}$, where $\boldsymbol{\lambda}_i$ is the vector of \emph{measured} interaction forces and torques exerted at the end-effector by the object, $\boldsymbol{\tau}_i$ denotes the vector of control inputs
(forces and torques), $\bs{{M}}_{q_i}(\bs{q}_i)$ is the inertial matrix, $\bs{{C}}_{q_i}({\bs{\zeta}}_i,\bs{q}_i)$ represents coriolis and centrifugal terms, $\bs{{D}}_{q_i}({\bs{\zeta}}_i,\bs{q}_i)$ models dissipative effects, $\bs{{g}}_i(\bs{q}_i)$ encapsulates the gravity and buoyancy effects and $\bs{d}_{q_i}(\bs{\zeta}_i,\bs{q}_i,t)$ is a bounded vector representing unmodeled friction, uncertainties and external disturbances. In view of \eqref{eq222}, we have:\vspace{-1mm}
\begingroup\makeatletter\def\f@size{9.5}\check@mathfonts
\def\maketag@@@#1{\hbox{\m@th\large\normalfont#1}}
\begin{align}
~~~~~~\dot{\bs{v}}_i=\bs{J}_i(\bs{q}_i)\dot{\bs{\zeta}}_i+\bs{J}^d_i(\bs{\zeta}_i,\bs{q}_i)\bs{\zeta}_i,~i\in\mathcal{K}\label{der_upsilon}
\end{align}\endgroup 
where $\bs{J}^d_i(\bs{\zeta}_i,\bs{q}_i)\in \mathbb{R}^{6\times(6+n)}$ represents the Jacobian derivative function (i.e., $\bs{J}^d_i(\bs{\zeta}_i,\bs{q}_i)\triangleq \dot{\bs{J}}_i(\bs{q}_i)$). Then, by employing the differential kinematics as well as \eqref{der_upsilon}, we obtain the transformed task space dynamics \cite{Siciliano-b1129198}:\vspace{-1mm}
\begingroup\makeatletter\def\f@size{9.5}\check@mathfonts
\def\maketag@@@#1{\hbox{\m@th\large\normalfont#1}}
\begin{align}
\!\bs{{M}}_{v_i}(\bs{q}_i)\dot{\bs{v}}_i\!+\!\bs{{C}}_{v_i}({\bs{\zeta}}_i,\bs{q}_i){\bs{v}}_i\!+&\bs{{D}}_{v_i}({\bs{\zeta}}_i,\bs{q}_i){\bs{v}}_i+\!\bs{{g}}_{v_i}(\bs{q}_i)+\nonumber\\ &\!\bs{d}_{v_i}(\bs{\zeta}_i,\bs{q}_i,t)=\!\boldsymbol{u}_i\!+\boldsymbol{\lambda}_i\label{UVMS_task_space}
\end{align}\endgroup
with the corresponding task space terms $\bs{M}_{v_i}\in \mathbb{R}^{6\times 6}$, $\bs{C}_{v_i}\in \mathbb{R}^{6\times 6}$, $\bs{D}_{v_i}\in \mathbb{R}^{6\times 6}$, $\bs{g}_{v_i}\in \mathbb{R}^{6}$, $\bs{d}_{v_i}\in \mathbb{R}^{6}$ and $\bs{u}_i$ that denotes the vector of task space generalized forces/torques.  Invoking the kinematic relations \eqref{eq4}-\eqref{dotv_obj}, we may express the aforementioned dynamics \eqref{UVMS_task_space} w.r.t the object's coordinates:\vspace{-2mm}
\begingroup\makeatletter\def\f@size{9.5}\check@mathfonts
\def\maketag@@@#1{\hbox{\m@th\large\normalfont#1}}
\begin{align}
\!\bs{{M}}_{i}(\bs{q}_i)\dot{\bs{v}}_O\!+\!\bs{{C}}_i({\bs{\zeta}}_i,\bs{q}_i){\bs{v}}_O\!+\!\bs{{D}}_i({\bs{\zeta}}_i,&\bs{q}_i){\bs{v}}_O\!+\!\bs{{g}}_i(\bs{q}_i)\!+\!\bs{d}_i(\bs{\zeta}_i,\bs{q}_i,t)\nonumber\\
&=\!\bs{J}_{iO}^\top\boldsymbol{u}_i\!+\bs{J}_{iO}^\top\boldsymbol{\lambda}_i\label{UVMS_object_fram}
\end{align}\endgroup
where:\vspace{-6mm}
\begingroup\makeatletter\def\f@size{9.5}\check@mathfonts
\def\maketag@@@#1{\hbox{\m@th\large\normalfont#1}}
\begin{gather*}
\bs{{M}}_{i}(\bs{q}_i)=\bs{J}_{iO}^\top\bs{{M}}_{v_i}(\bs{q}_i)\bs{J}_{iO}\\
\bs{{C}}_i({\bs{\zeta}}_i,\bs{q}_i)=\bs{J}_{iO}^\top\big[ \bs{{C}}_{v_i}({\bs{\zeta}}_i,\bs{q}_i) \bs{J}_{iO}+ \bs{{M}}_{v_i}(\bs{q}_i) \dot{\bs{J}}_{iO}  \big]\\
\bs{{D}}_i({\bs{\zeta}}_i,\bs{q}_i)=\bs{J}_{iO}^\top \bs{{D}}_{v_i}({\bs{\zeta}}_i,\bs{q}_i) \bs{J}_{iO}\\
\bs{{g}}_i(\bs{q}_i)=\bs{J}_{iO}^\top \bs{{g}}_{v_i}(\bs{q}_i),\quad~\bs{d}_i(\bs{\zeta}_i,\bs{q}_i,t)=\bs{J}_{iO}^\top\bs{d}_{v_i}(\bs{\zeta}_i,\bs{q}_i,t)
\end{gather*}\endgroup
Now, the following common properties will be employed in the analysis.
\begin{Property}\label{Property1} 
	The matrix $\bs{M}_i(\bs{q}_i),~i\in\mathcal{K}$ is positive definite and the matrix $\dot{\bs{M}}_i(\bs{q}_i)-2\bs{{C}}_i({\bs{\zeta}}_i,\bs{q}_i),~ i\in\mathcal{K}$ is skew-symmetric.%\vspace{-2mm}
\end{Property}
\begin{Property}\label{Property2} 
	The uncertainty of the UVMS model appears linearly in the dynamics \eqref{UVMS_object_fram} in terms of an unknown but constant parameter vector $\boldsymbol{\theta}_i \in \mathbb{R}^{q_i},~ i\in\mathcal{K}$ in the following way\cite{Slotine1987595,Tatlicioglu2009873}:\vspace{-1mm}
	\begingroup\makeatletter\def\f@size{8.5}\check@mathfonts
	\def\maketag@@@#1{\hbox{\m@th\large\normalfont#1}}
	\begin{gather*}
	\bs{{M}}\!_i(\bs{a}_i\!){\bs{d}}_i\!+\!\bs{{C}}\!_i({\bs{a}}_i,\bs{b}_i\!){\bs{c}}_i\!+\!\bs{{D}}\!_i({\bs{a}}_i,\bs{b}_i\!){\bs{c}}_i\!+\!\bs{{g}}\!_i(\bs{a}_i)\!=\!\boldsymbol{\Omega}\!_i(\bs{a}_i,{\bs{b}}_i,{\bs{c}}_i,{\bs{d}}_i\!)\boldsymbol{\theta}_i,\!~i\!\in\!\mathcal{K}
	\end{gather*}\endgroup
	where $\boldsymbol{\Omega}_i(\bs{a}_i,{\bs{b}}_i,{\bs{c}}_i,{\bs{d}}_i)\in\mathbb{R}^{6\times q_i},~i\in\mathcal{K}$ is a regressor matrix of known functions of $\bs{a}_i,{\bs{b}}_i,{\bs{c}}_i,{\bs{d}}_i \in \mathbb{R}^{6}$  independent of $\boldsymbol{\theta}_i$.
\end{Property}
Now, we introduce the following assumption regarding the unmodeled dynamics/external disturbances. 
\begin{assumption}\label{assum_distur}
	There exist unknown constant vector $\bs{\theta}_{d,i}$ and known bounded functions $\bs{\Delta}_i\in\mathbb{R}^{6\times q_i}$, such that\vspace{-1mm}
	\begingroup\makeatletter\def\f@size{9.5}\check@mathfonts
	\def\maketag@@@#1{\hbox{\m@th\large\normalfont#1}}
	\begin{align*}
	\bs{d}_i(\bs{\zeta}_i,\bs{q}_i,t)=\bs{\Delta}_i(\bs{\zeta}_i,\bs{q}_i,t)\bs{\theta}_{d,i},~i\in\mathcal{K}.
	\end{align*}\endgroup
\end{assumption}\vspace{-0mm}
\textbf{Object Dynamics:} Without any loss of generality, we consider the following second order dynamics for the object, which can be derived based on the Newton-Euler formulations:\vspace{-1mm}
\begingroup\makeatletter\def\f@size{9.0}\check@mathfonts
\def\maketag@@@#1{\hbox{\m@th\large\normalfont#1}}
\begin{subequations}
	\begin{align}
	&\!\!\!\!\!\dot{\bs{x}}_O={\bs{J}_O({\bs{\eta}}_{2,O})}\bs{v}_O\label{obj_dyn_a}\\
	&\!\!\!\!\!\!\bs{M}\!_O(\bs{x}_O)\dot{\bs{v}}_O\!+\!\bs{C}\!_O(\dot{\bs{x}}_O,\bs{x}_O)\bs{v}_O\!+\!\bs{D}\!_O(\dot{\bs{x}}_O,\bs{x}_O)\bs{v}_O\!\!+\!\!\bs{g}\!_O\!\!=\!\!\bs{\lambda}\!_O\!\!+\!\!\bs{\lambda}\!_{e}\!\label{obj_dyn_b}\end{align}
\end{subequations}\endgroup
where $\bs{M}_O(\bs{x}_O)$ is the positive definite inertia matrix, $\bs{C}_O(\dot{\bs{x}}_O,\bs{x}_O)$ is the Coriolis matrix, $\bs{g}_O$ is the vector of gravity and buoyancy effects, $\bs{D}_O(\dot{\bs{x}}_O,\bs{x}_O)$ models dissipative effects,  $\bs{\lambda}_O$ is the vector of generalized forces acting on the object's center of mass, $\bs{\lambda}_{e}$ is a vector representing uncertainties and external disturbances and ${\bs{J}_O({\bs{\eta}}_{2,O})}=\text{diag}\{\bs{I}_3,{\bs{J}'_O({\bs{\eta}}_{2,O})}\}$ is the object representation Jacobian with:\vspace{-1mm}
\begingroup\makeatletter\def\f@size{9.5}\check@mathfonts
\def\maketag@@@#1{\hbox{\m@th\large\normalfont#1}}
\begin{gather}\label{object_Euler-vel}
\bs{J}'_O({\bs{\eta}}_{2,O})=\begin{bmatrix}
1 &  \sin(\phi_O)\tan(\theta_O)&cos(\phi_O)\tan(\theta_O)\\
0&\cos(\phi_O) & -\sin(\theta_O)\\
0&\frac{\sin(\phi_O)}{\cos(\theta_O)}&\frac{cos(\phi_O)}{\cos(\theta_O)}
\end{bmatrix}.\end{gather}\endgroup
Moreover, The kineto-statics duality along with the grasp rigidity suggest that the force $\bs{\lambda}_O$ acting on the object's center of mass and the generalized forces $\bs{\lambda}_i,~i\in \mathcal{K}$, exerted by the UVMSs at the grasping points, are related through:\vspace{-1mm}
\begingroup\makeatletter\def\f@size{9.5}\check@mathfonts
\def\maketag@@@#1{\hbox{\m@th\large\normalfont#1}}
\begin{align}\label{object_grasp_matrix}
\bs{\lambda}_O=\bs{G}^\top\bs{\lambda}
\end{align}\endgroup
where:\vspace{-1mm}
\begingroup\makeatletter\def\f@size{8.5}\check@mathfonts
\def\maketag@@@#1{\hbox{\m@th\large\normalfont#1}}
\begin{align}\label{grasp matrix}
~~~~~~\bs{G}\!=\!\Big[[\bs{J}_{LO}]^\top\!\!,[\bs{J}_{F_1O}]^\top\!\!,\ldots,[\bs{J}_{F_NO}]^\top\!\Big]^\top\!\!\!\in \mathbb{R}^{6(N+1)\times 6 }
\end{align}\endgroup
is the full column-rank grasp matrix and $\bs{\lambda}=[\bs{\lambda_L}^\top,\bs{\lambda_{F_1}}^\top,\ldots,\bs{\lambda_{F_N}}^\top]^\top$ is the vector of the overall interaction forces and torques.  \vspace{-1mm}
\subsection{Description of the Workspace}\vspace{0mm}
Consider the team of $N+1$ UVMSs operating in a bounded workspace $\mathcal{W} \subseteq \mathbb{R}^3$ with boundary $\partial\mathcal{W}$.  Without any loss of the generality, the obstacles, the robots as well as the workspace are all modeled by spheres (i.e., we adopt the spherical world representation \cite{Koditschek1990412}). In this spirit, let $\mathcal{B}(\bs{x}_O, r_0)$ be a closed ball that covers the volume of the object and has radius $r_0$. We also define the closed balls $\mathcal{B}(\boldsymbol{p}_i, \bar{r}), i\in \mathcal{K}$, centered at the end-effector of each UVMS that cover the robot volume for all possible configurations. Furthermore, we define  a ball area $\mathcal{B}(\bs{x}_O,R)$ located at $\bs{x}_O$ with radius $R=\bar{r}+r_o$ that includes the whole volume of the robotic team and the object. Finally, the $\mathcal{M}$ static obstacles are defined as closed spheres described by $\pi_m=\mathcal{B}(\bs{p}_{\pi_m},r_{\pi_m}),~m\in\{1,\ldots, \mathcal{M}\}$, where $\bs{p}_{\pi_m}\in \mathbb{R}^3$ is the center and the $r_{\pi_m}$ the radius of the obstacle $\pi_m$.  Additionally, based on the property of spherical world representation \cite{Koditschek1990412}, each pair of obstacles $m,m'\in \{1,\ldots,\mathcal{M}\}$ are disjoint in a such a way that the whole team of UVMSs including the object can pass through the free space between them. Therefore, there exists a feasible trajectory $\bs{x}_O(t)$ for the whole team that connects the initial configuration $\bs{x}_O(t_0)$ with $\bs{x}^d_O$. \vspace{-2mm}
\section{Control Methodology}\vspace{-1mm}
We assume that the leading UVMS is aware of both the desired configuration of the object as well as of the obstacles position in the workspace. Towards this direction, and based on the property of spherical world representation \cite{Koditschek1990412} we assume that there is a safe trajectory within the workspace which is known only to the leader:\vspace{-2mm}
\subsection{Safe Navigation}\vspace{-1mm}
The desired/feasible object trajectory within the workspace $\mathcal{W}$ can be generated based on the Navigation Functions concept originally proposed by Rimon and Koditschek in \cite{Koditschek1990412}, as follows:\vspace{-1mm}
\begingroup\makeatletter\def\f@size{9.5}\check@mathfonts
\def\maketag@@@#1{\hbox{\m@th\large\normalfont#1}} 	
\begin{align}
~~~~~~~~~~~\phi_O(\bs{x}_O;\bs{x}^d_O)=\frac{\gamma(\bs{x}_O-\bs{x}^d_O)}{[\gamma^k(\bs{x}_O-\bs{x}^d_O)+\beta(\bs{x}_O)]^{\frac{1}{k}}}\label{eq8}
\end{align}\endgroup

\vspace{-2.5mm}
\noindent where $\phi_O:\xrightarrow{\mathcal{W}-\overset{\mathcal{M}}{\underset{m=1}{\cap}}\mathcal{B}(\bs{p}_{\pi_m},r_{\pi_m})}[0,1)$ denotes the potential that derives a safe motion vector field within the free space $\mathcal{W}-\overset{\mathcal{M}}{\underset{m=1}{\cap}}\mathcal{B}(\bs{p}_{\pi_m},r_{\pi_m})$. Notice that $k>1$ is a design constant, $\gamma(\bs{x}_O-\bs{x}^d_O)>0$ with $\gamma(\bs{0})=0$ represents the attractive potential field to the goal configuration $\bs{x}^d_O$ and $\beta(\bs{x}_O)>0$ with:\vspace{-1mm}
\begingroup\makeatletter\def\f@size{9.5}\check@mathfonts
\def\maketag@@@#1{\hbox{\m@th\large\normalfont#1}}
\begin{equation*}
\lim\limits_{\bs{x}_O\rightarrow \begingroup\makeatletter\def\f@size{6}\check@mathfonts	
	\arraycolsep=0.4pt\def\arraystretch{0.6} \Bigg\{
	\begin{array}{ccc}
	\text{Boundary} \\
	\text{Obstacles}
	\end{array} 
	\endgroup}   \beta(\bs{x}_O)=0
\end{equation*}\endgroup

\vspace{-3.0mm}
\noindent represents the repulsive potential field by the workspace boundary and the obstacle regions. In that respect, it was proven in \cite{Koditschek1990412} that $\phi_O(\bs{x}_O;\bs{x}^d_O)$ has a global minimum at $\bs{x}^d_O$ and no other local minima for sufficiently large $k$. Thus, a feasible path that leads from any initial obstacle-free configuration to the desired configuration might be generated by following the negated gradient of $\phi_O(\bs{x}_O;\bs{x}^d_O)$. Consequently, the desired velocity profile at the leader's side is designed as follows:\vspace{-1mm}
\begingroup\makeatletter\def\f@size{9.5}\check@mathfonts
\def\maketag@@@#1{\hbox{\m@th\large\normalfont#1}}
\begin{align}
~~~~~~~~{\bs{v}}^d_{O_L}(t)\!=\!-K_{NF}{\bs{J}^{-1}_O({\bs{\eta}}_{2,O})}\nabla_{\bs{x}_O}\phi_O(\bs{x}_{O_L}(t),\bs{x}^d_O)\label{eq9}
\end{align}\endgroup
where $K_{NF}>0$ is a positive gain.  Given the initial configuration, the leading UVMS may calculate the desired trajectory and velocity profile denoted by  $\bs{x}^d_{O_L}(t)$ and $\bs{v}^d_{O_L}(t)$ respectively, by propagating the model $\dot{\bs{x}}^d_{O_L}(t)=\bs{v}^d_{O_L}(t)$. \vspace{-1mm}
\subsection{Control Design}
First, we introduce  the load sharing coefficients $c_i,~i\in\mathcal{K}$ that are subject to the following design constraints: $
c_i\in (0,1), \forall i\in\mathcal{K}~\text{and} ~\sum_{i\in\mathcal{K}} c_i=1$. Without any loss of generality and to simplify the analysis we select:\vspace{-2mm}
\begingroup\makeatletter\def\f@size{9.5}\check@mathfonts
\def\maketag@@@#1{\hbox{\m@th\large\normalfont#1}}
\begin{align}
c_i=\frac{1}{N+1},\quad~i\in\mathcal{K}.\label{C_I}
\end{align}\endgroup
In view of the object dynamics \eqref{obj_dyn_b}, it can be concluded that the vector of external disturbances $\bs{\lambda}_{e}$ is unknown. Thus, in order to design the impedance control scheme, each UVMS should estimate the aforementioned vector in a distributed way (since explicit communication among UVMSs is not permitted). 
%Moreover, based on the object dynamics \eqref{obj_dyn_b}, the vector of external disturbances is impossible to be estimated in a decentralized way by relying only on local measurements (i.e., local force torque measurements at UVMS's end effector), since it depends on the applying force from all member of the UVMS teams on the object. 
Therefore, an online estimation method based on the object momentum concept \cite{de2005sensorless} is given in the sequel.  Based on the load coefficients \eqref{C_I} the object dynamics of \eqref{obj_dyn_b} can be rewritten as: \vspace{-2mm}
\begingroup\makeatletter\def\f@size{9.5}\check@mathfonts
\def\maketag@@@#1{\hbox{\m@th\large\normalfont#1}}
\begin{align}
&\sum_{i\in\mathcal{K}}\!\!\big\{\!\bs{M}_{O_i}\!(\bs{x}_O)\dot{\bs{v}}_O\!+\!\bs{C}_{O_i}\!(\dot{\bs{x}}_O,\bs{x}_O)\bs{v}_O\!+\!\bs{D}_{O_i}\!(\dot{\bs{x}}_O,\bs{x}_O)\bs{v}_O\!+\!\bs{g}_{O_i}\!\big\}\nonumber\\
&\qquad\qquad\qquad\qquad=\sum_{i\in\mathcal{K}}\!\bs{J}\!_{iO}^\top\bs{\lambda}_i+\sum_{i\in\mathcal{K}}\!\bs{\lambda}_{e_i}\label{obj_ci}
\end{align}\endgroup
where $\bs{M}_{O_i}=c_i\bs{M}_O$, $\bs{C}_{O_i}=c_i\bs{C}_O$, $\bs{D}_{O_i}=c_i\bs{D}_O$ , $\bs{g}_{O_i}=c_i\bs{g}_O$ and $\bs{\lambda}_{e_i}=c_i\bs{\lambda}_e$. In order to estimate  $\bs{\lambda}_{e_i}$ locally we define the object equivalent momentum \cite{de2005sensorless} $\bs{\mu}_i=\bs{M}_{O_i}{\bs{v}}_O$ and the vector $\bs{\zeta}_i(t)\in \mathbb{R}^6$ as:\vspace{-2mm}
\begingroup\makeatletter\def\f@size{9.5}\check@mathfonts
\def\maketag@@@#1{\hbox{\m@th\large\normalfont#1}}
\begin{align}
\!\!\bs{\zeta}_i(t)\!=\!\bs{K}\!_\mu\Bigg(\!\!\bs{\mu}_i(t)\!+\!\!\int_{t_0}^{t}\!\!\!\Big( \!\bs{C}_{O_i}\!\bs{v}_O\!+\!\bs{D}_{O_i}\!\bs{v}_O\!+\!\bs{g}_{O_i}\! -\!\bs{\zeta}_i(d\tau)\Big)d\tau\! \Bigg)\!\!\!\label{momen_obj}
\end{align}\endgroup
whose time derivative is given by:\vspace{-2mm}
\begingroup\makeatletter\def\f@size{9.5}\check@mathfonts
\def\maketag@@@#1{\hbox{\m@th\large\normalfont#1}}
\begin{align}
\dot{\bs{\zeta}}_i(t)\!=-\!\bs{K}\!_\mu\bs{\zeta}_i(t)+c_i\bs{K}\!_\mu\Big( \sum_{i\in\mathcal{K}}\!\bs{J}\!_{iO}^\top\bs{\lambda}_i +  \bs{\lambda}_e\Big)\label{lowpass}
\end{align}\endgroup
where $\bs{K}_\mu$ is a positive definite matrix gain. Notice that for a sufficiently large gain matrix $\bs{K}_\mu$ of the low pass filter \eqref{lowpass}, we obtain:\vspace{-4mm}
\begingroup\makeatletter\def\f@size{9.5}\check@mathfonts
\def\maketag@@@#1{\hbox{\m@th\large\normalfont#1}}
\begin{align}
\bs{\zeta}_i(t)\! \approx c_i\Big(\sum_{i\in\mathcal{K}}\!\bs{J}\!_{iO}^\top\bs{\lambda}_i +  \bs{\lambda}_e\Big)
\end{align}\endgroup
which intuitively means that $\bs{\zeta}_i(t)$ represents the effect of overall external forces exerted on the object (i.e., external disturbances and the forces exerted by all the UVMSs on the object). Consequently, an estimation of $\bs{\lambda}_{e_i}=c_i\bs{\lambda}_e$ can be given by:\vspace{-5mm}
\begingroup\makeatletter\def\f@size{9.5}\check@mathfonts
\def\maketag@@@#1{\hbox{\m@th\large\normalfont#1}}
\begin{align}
{\bs{\lambda}}_{e_i}  \approx \bs{\zeta}_i(t)-\bs{J}\!_{iO}^\top\bs{\lambda}_i ,~ i\in\mathcal{K}.\label{external_estimate}
\end{align}\endgroup
Now let us assume that each UVMS is expected to exert the following desired force/torque on the object:\vspace{-2mm}
\begingroup\makeatletter\def\f@size{9.5}\check@mathfonts
\def\maketag@@@#1{\hbox{\m@th\large\normalfont#1}}
\begin{align}
\!\!\bs{\lambda}^d_i \!\!=\! \bs{\lambda}^d\!\!_{\!int,i}\!-\!\bs{J}_{iO}^{-\!\top}\!(\bs{M}\!_{O_i}{\bs{y}}^{cmd}_{i}\!\!+\!\bs{C}\!_{O_i}\!\bs{v}_O\!+\!\bs{D}\!_{O_i}\!\bs{v}_O\!+\!\bs{g}_{O_i}\!\!-\!\!\bs{\lambda}_{e_i}\!)\!\!\!\label{desired_wrench}
\end{align}\endgroup
where $\bs{\lambda}^d_{int,i}$  denotes the desired internal forces and ${\bs{y}}^{cmd}_{i}$ is a pre-designed input given by:\vspace{-1mm}
\begingroup\makeatletter\def\f@size{9.5}\check@mathfonts
\def\maketag@@@#1{\hbox{\m@th\large\normalfont#1}}
\begin{align}
{\bs{y}}^{cmd}_{i}\!=\! \dot{\bs{v}}^d_{O_i}\!+\!\bs{M}_{d_O}^{-1}\Big[ \!-\!\bs{D}_{d_O} \tilde{\bs{v}}_O\!-\!\bs{K}_{d_O} \tilde{\bs{e}}_O\Big]
\end{align}\endgroup
where  $\bs{M}_{d_O}$, $\bs{D}_{d_O}$ and $\bs{K}_{d_O}$ are the desired inertia, damping and stiffness matrices for the object dynamics, $\tilde{\bs{v}}_O(t)={\bs{v}}_{O}-{\bs{v}}^d_{O_i}$ denotes the velocity error and $\tilde{\bs{e}}_O$ is the object pose error, defined as:\vspace{-1mm}
\begingroup\makeatletter\def\f@size{9.5}\check@mathfonts
\def\maketag@@@#1{\hbox{\m@th\large\normalfont#1}}
\begin{align}
\tilde{\bs{e}}_O=\begin{bmatrix}
\bs{\eta}_{1,O}-\bs{\eta}^d_{1,O}\\
\tilde{\bs{\epsilon}}_O
\end{bmatrix}
\end{align}\endgroup
where\vspace{-6mm}
\begingroup\makeatletter\def\f@size{9.5}\check@mathfonts
\def\maketag@@@#1{\hbox{\m@th\large\normalfont#1}}
\begin{align}
~~~~~\tilde{\bs{\epsilon}}_O\!=\!\frac{1}{2}\Big(\bs{n}_O\!\times\!\bs{n}^d_O\!+\!\bs{o}_O\!\times\!\bs{o}^d_O\!+\!\bs{\alpha}_O\!\times\!\bs{\alpha}^d_O\Big)\label{err_4}~\!\in\mathbb{R}^3
\end{align}\endgroup
is the orientation error expressed in the outer product formulation \cite{caccavale1998resolved}. In view of \eqref{obj_ci}, it can be concluded that if all robots cooperatively apply the desired wrench vector \eqref{desired_wrench} to the object, then\vspace{-1mm}
\begingroup\makeatletter\def\f@size{9.5}\check@mathfonts
\def\maketag@@@#1{\hbox{\m@th\large\normalfont#1}}
\begin{align}
~~~~~~\bs{M}_{d_O}\dot{\tilde{\bs{v}}}_O+\bs{D}_{d_O}\tilde{\bs{v}}_O+\bs{K}_{d_O}\tilde{\bs{e}}_O=\bs{0}
\end{align}\endgroup
which intuitively means that the aforementioned selection of $\bs{\lambda}^d_i$ cancels the object's nonlinearities, ensures adequate internal forces via $\bs{\lambda}^d_{int,i}$ and achieves the desired dynamics of the object. Thus, the control objective for each UVMS $i\in\mathcal{K}$ is to enforce $\lim_{t\rightarrow\infty}\bs{w}_i(t)=0$, where the error signal $\bs{w}(t)$ is constructed as:\vspace{-1mm}
\begingroup\makeatletter\def\f@size{9.5}\check@mathfonts
\def\maketag@@@#1{\hbox{\m@th\large\normalfont#1}}
\begin{align}
\bs{w}_i(t)\!=\!\bs{M}_{d}\dot{\tilde{\bs{v}}}_O\!+\!\bs{D}_{d}\tilde{\bs{v}}_O\!+\!\bs{K}_{d}\tilde{\bs{e}}_O\!-\!\bs{J}\!_{iO}^\top\bs{\lambda}^d_i,~i\in\mathcal{K}
\end{align}\endgroup
where $\bs{M}_{d}$, $\bs{D}_{d}$ and $\bs{K}_{d}$ are the desired inertia, damping and stiffness matrices for the robot dynamics. Thus, we get an augmented impedance error:\vspace{-2mm}
\begingroup\makeatletter\def\f@size{9.5}\check@mathfonts
\def\maketag@@@#1{\hbox{\m@th\large\normalfont#1}}
\begin{align}
\tilde{\bs{w}}_i=\bs{K}_f{\bs{w}}_i=\dot{\tilde{\bs{v}}}_O+\bs{K}_{g}\tilde{\bs{v}}_O+\bs{K}_{p}\tilde{\bs{e}}_O-\bs{K}_f\bs{J}\!_{iO}^\top\bs{\lambda}^d_i\label{eq32}
\end{align}\endgroup
where $\bs{K}_f=\bs{M}_{d}^{-1}$, $\bs{K}_{g}=\bs{K}_f\bs{D}_{d}$, and $\bs{K}_{p}=\bs{K}_f\bs{K}_{d}$. We also choose two positive-definite matrices $\bs{F}$ and $\bs{Y}$ such that:\vspace{-2mm}
\begingroup\makeatletter\def\f@size{9.5}\check@mathfonts
\def\maketag@@@#1{\hbox{\m@th\large\normalfont#1}}
\begin{gather*}
\bs{F}+\bs{Y}=\bs{K}_g,~\text{and}~ \dot{\bs{F}}+\bs{Y}\bs{F}=\bs{K}_{p}
\end{gather*}\endgroup 

\vspace{-2mm}
\noindent and define the filtered force/torque measurement:\vspace{-1mm}
\begingroup\makeatletter\def\f@size{9.5}\check@mathfonts
\def\maketag@@@#1{\hbox{\m@th\large\normalfont#1}}
\begin{align}
\dot{\bs{f}}_i+\bs{Y}\bs{f}_i=\bs{K}_f\bs{J}\!_{iO}^\top\bs{\lambda}^d_i,~i\in\mathcal{K}.
\end{align}\endgroup

\vspace{-2mm}
\noindent Thus, we may rewrite \eqref{eq32} as:\vspace{-1mm}
\begingroup\makeatletter\def\f@size{9.5}\check@mathfonts
\def\maketag@@@#1{\hbox{\m@th\large\normalfont#1}}
\begin{align}
\tilde{\bs{w}}_i=\dot{\tilde{\bs{v}}}_O+(\bs{F}+\bs{Y})\tilde{\bs{v}}_O+(\dot{\bs{F}}+\bs{Y}\bs{F})\tilde{\bs{e}}_O-\dot{\bs{f}}_i-\bs{Y}\bs{f}_i.\label{eq34}
\end{align}\endgroup

\vspace{-2mm}
\noindent Now we define the auxiliary variable $\bs{z}_i,~i\in\mathcal{K}$ as:\vspace{-1mm}
\begingroup\makeatletter\def\f@size{9.5}\check@mathfonts
\def\maketag@@@#1{\hbox{\m@th\large\normalfont#1}}
\begin{align}
\bs{z}_i=\tilde{\bs{v}}_O+\dot{\bs{F}}\tilde{\bs{e}}_O-\bs{f}_i,~i\in\mathcal{K}.\label{eq35}
\end{align}\endgroup

\vspace{-2mm}
\noindent Hence, the augmented impedance error becomes:\vspace{-2mm}
\begingroup\makeatletter\def\f@size{9.5}\check@mathfonts
\def\maketag@@@#1{\hbox{\m@th\large\normalfont#1}}
\begin{align}
\tilde{\bs{w}}_i=\dot{\bs{z}}_i+\bs{Y}\bs{z},~i\in\mathcal{K}\label{eq355}
\end{align}\endgroup

\vspace{-2mm}
\noindent which represents a stable low pass filter. Therefore, if we achieve $\lim_{t\rightarrow\infty}\bs{z}_i(t)=0$, then the initial control objective is readily met, i.e, $\lim_{t\rightarrow\infty}\bs{w}_i(t)=0$. In this respect, let us define the augmented state variable:\vspace{-2.5mm}
\begingroup\makeatletter\def\f@size{9.5}\check@mathfonts
\def\maketag@@@#1{\hbox{\m@th\large\normalfont#1}}
\begin{align}
\bs{v}_{O_i}^r=\bs{v}_{O_i}^d-\bs{F}\tilde{\bs{e}}_O+\bs{f}_i,~i\in\mathcal{K}\label{eq37}
\end{align}\endgroup 
Hence, \eqref{eq35} and \eqref{eq37} immediately result in:\vspace{-3mm}
\begingroup\makeatletter\def\f@size{9.5}\check@mathfonts
\def\maketag@@@#1{\hbox{\m@th\large\normalfont#1}}
\begin{align}
\bs{z}_i=\bs{v}_O-\bs{v}_{O_i}^r,~i\in\mathcal{K}\label{eq38}
\end{align}\endgroup
from which the dynamics \eqref{UVMS_object_fram} becomes:\vspace{-2mm}
\begingroup\makeatletter\def\f@size{8.5}\check@mathfonts
\def\maketag@@@#1{\hbox{\m@th\large\normalfont#1}}
\begin{align*}
\!\bs{{M}}\!_{i}\dot{\bs{z}_i}\!+\!\bs{{C}}\!_i{\bs{z}_i}\!+\!\bs{{D}}_i{\bs{z}}\!=\!\bs{J}\!_{iO}^\top\boldsymbol{u}_i\!+\!\bs{J}_{iO}^\top\boldsymbol{\lambda}_i\!-\!\Big[\! \!\bs{{M}}\!_{i}\dot{\bs{v}}_{O_i}^r\!\!+\!\bs{{C}}\!_i\bs{v}_{O_i}^r\!\!+\!\bs{{D}}\!_i\bs{v}_{O_i}^r\! \!+\!\bs{{g}}_i\!\!+\!\bs{d}_i\! \Big].
\end{align*}\endgroup
Invoking Property $2$ and Assumption \ref{assum_distur} , we arrive at the open loop dynamics:\vspace{-2mm}
\begingroup\makeatletter\def\f@size{9.5}\check@mathfonts
\def\maketag@@@#1{\hbox{\m@th\large\normalfont#1}}
\begin{align}
\!\bs{{M}}_{i}\dot{\bs{z}_i}\!+\!\bs{{C}}_i{\bs{z}_i}\!+&\!\bs{{D}}_i{\bs{z}_i}\!=\!\bs{J}_{iO}^\top\boldsymbol{u}_i\!+\bs{J}_{iO}^\top\boldsymbol{\lambda}_i-\bs{\Delta}_i(\bs{\zeta}_i,\bs{q}_i,t){\bs{\theta}}_{d,i}\nonumber\\&-\boldsymbol{\Omega}_i(\bs{q}_i,{\bs{\zeta}}_i,\bs{v}_{O_i}^r,\dot{\bs{v}}_{O_i}^r){\boldsymbol{\theta}}_i,~i\in\mathcal{K}.\label{eq39}
\end{align}\endgroup  
Therefore, we design the following impedance control scheme:\vspace{-2mm}
\begingroup\makeatletter\def\f@size{9.5}\check@mathfonts
\def\maketag@@@#1{\hbox{\m@th\large\normalfont#1}}
\begin{align}
\bs{u}_i=-\boldsymbol{\lambda}_i+\bs{J}_{iO}^{-\top}\!&\Big[\boldsymbol{\Omega}_i(\bs{q}_i,{\bs{\zeta}}_i,\bs{v}_{O_i}^r,\dot{\bs{v}}_{O_i}^r)\hat{\boldsymbol{\theta}}_i\!\nonumber\\
&+\! \bs{\Delta}_i(\bs{\zeta}_i,\bs{q}_i,t)\hat{\bs{\theta}}_{d,i}-\bs{K}\bs{z}_i \Big],~i\in\mathcal{K}\label{eq10a}
\end{align}\endgroup
where $\bs{K}>0$ is a positive definite gain matrix and $\hat{\boldsymbol{\theta}}_i$, $\hat{\boldsymbol{\theta}}_{d,i}$ denote the estimates of the unknown parameters $\boldsymbol{\theta}_i$,  $\boldsymbol{\theta}_{d,i}$ respectively, provided by the update laws:\vspace{-2mm}
\begingroup\makeatletter\def\f@size{9.5}\check@mathfonts
\def\maketag@@@#1{\hbox{\m@th\large\normalfont#1}}
\begin{align}
\dot{\hat{\boldsymbol{\theta}}}_i = -\bs{\Gamma}_i &\bs{\Omega}_i(\bs{q}_i,{\bs{\zeta}}_i,{\bs{v}}^r_{O_i},\dot{\bs{v}}^r_{O_i}){\bs{z}}_{},~ \boldsymbol{\Gamma}_i>0\label{eq10b}\\
\dot{\hat{\boldsymbol{\theta}}}_{d,i} = -\boldsymbol{\Gamma}&_{d_i} \bs{\Delta}_i(\bs{\zeta}_i,\bs{q}_i,t){\bs{z}}_{},~ \boldsymbol{\Gamma}_{d_i}>0\label{eq10c}
\end{align}\endgroup
with  $\boldsymbol{\Gamma}_i$, $\boldsymbol{\Gamma}_{d_i}$ positive diagonal gain matrices.
\begin{th1} \label{th1}
	Consider $N+1$ UVMSs that operate in a constrained workspace $\mathcal{W}$, with dynamics \eqref{UVMS_object_fram} obeying Properties \ref{Property1}--\ref{Property2} and grasp rigidly a common object. The control scheme \eqref{eq10a} with adaptive laws \eqref{eq10b}-\eqref{eq10c} guarantees $\lim_{t\rightarrow\infty}\bs{w}_i(t)=0$ as well as the boundedness of all signals in the closed loop system.
\end{th1}

\textbf{Proof:}	Considering the following Lyapunov function candidate:\vspace{-4mm}
\begingroup\makeatletter\def\f@size{9.5}\check@mathfonts
\def\maketag@@@#1{\hbox{\m@th\large\normalfont#1}}
\begin{align*}
~~~~~~~\bs{V}\!=\!\sum_{i\in\mathcal{K}}\!\!\frac{1}{2}\bs{z}_i^\top\!\bs{M}_i\bs{z}_i\!+\!\sum_{i\in\mathcal{K}}\!\!\frac{1}{2}\tilde{\bs{\theta}}_i^\top\!\bs{\Gamma}_i^{-1}\tilde{\bs{\theta}}_i\!+\!\sum_{i\in\mathcal{K}}\!\!\frac{1}{2}\tilde{\bs{\theta}}_{d_i}^\top\!\bs{\Gamma}_{d_i}^{-1}\tilde{\bs{\theta}}_{d_i}
\end{align*}\endgroup
where $\tilde{\bs{\theta}}_i=\hat{\bs{\theta}}_i-{\bs{\theta}}_i$ and $\tilde{\bs{\theta}}_{d_i}=\hat{\bs{\theta}}_{d_i}-{\bs{\theta}}_{d_i}$ denote the parametric errors, differentiating with respect to time and invoking Property $1$ and substituting the control scheme \eqref{eq10a}-\eqref{eq10c}, we get:\vspace{-4mm}
\begingroup\makeatletter\def\f@size{9.5}\check@mathfonts
\def\maketag@@@#1{\hbox{\m@th\large\normalfont#1}}
\begin{align}
\dot{\bs{V}}=\sum_{i\in\mathcal{K}}-\bs{z}_i^\top\bs{K}\bs{z}_i-\bs{z}_i^\top\bs{D}_i\bs{z}_i\leq\bs{0}\label{eq43}
\end{align}\endgroup
Hence,  we conclude that $\bs{z}_i$, $\tilde{\bs{\theta}}_i$, $\tilde{\bs{\theta}}_{d_i} \in L_\infty$. Moreover, from the definition of $\bs{z}_i$ in \eqref{eq38} , we also deduce that $\bs{x}_O, \bs{v}_O\in L_\infty$, and consequently  $\bs{v}_{O_i}^r, \dot{\bs{v}}_{O_i}^r\in L_\infty$. Furthermore, employing \eqref{eq39} we arrive at $\dot{\bs{z}}\in L_\infty$. Therefore, integrating both sides of \eqref{eq43} leads to:\vspace{-3mm}
\begingroup\makeatletter\def\f@size{9.5}\check@mathfonts
\def\maketag@@@#1{\hbox{\m@th\large\normalfont#1}}
\begin{equation}
\bs{V}(t)\!-\!\bs{V}(0)\leq\sum_{i\in\mathcal{K}}\!\int_{0}^t \!\!\Big(\!-\!\bs{z}_i^\top(\tau)\bs{K}\bs{z}_i\!-\!\bs{z}_i^\top\bs{D}_i\bs{z}_i(\tau)\Big)d\tau\!\label{eq44}
\end{equation}\endgroup
Thus, $\int_{0}^t\big(\!-\!\bs{z}_i^\top(\tau)\bs{K}\bs{z}_i\!-\!\bs{z}_i^\top\bs{D}_i\bs{z}_i(\tau)\big)d\tau$ is bounded, which results in $\bs{z}_i\in L_2$. Finally, applying Barbalat's Lemma and invoking \eqref{eq355} we get $\lim_{t\rightarrow\infty}\bs{w}_i(t)\rightarrow\bs{0},~\forall i\in\mathcal{K}$  which completes the proof.\vspace{-2mm}
\subsection{Estimation Scheme}
It should be noticed that the followers are not aware of either the object's desired configuration $\bs{x}^d_O$ or the obstacles' position in the workspace. However, the followers will estimate the object's desired trajectory profile by $\hat{\bs{x}}^{d_i}_O(t)~i\in \mathcal{N}$, via their own state measurements by adopting a prescribed performance estimator. Hence, let us define the error:\vspace{-2mm}
\begingroup\makeatletter\def\f@size{9.5}\check@mathfonts
\def\maketag@@@#1{\hbox{\m@th\large\normalfont#1}}
\begin{align}\bs{e}_i(t)=\bs{x}_O(t)-\hat{\bs{x}}^{d_i}_O(t)\in\mathbb{R}^6,~i\in\mathcal{N}.\end{align}\endgroup
The expression of prescribed performance for each element of $\bs{e}_i(t)=[e_{i1}(t),\ldots,e_{i6}(t)]^\top,~i\in\mathcal{N}$ is given by the following inequalities:\vspace{-2mm}
\begingroup\makeatletter\def\f@size{9.5}\check@mathfonts
\def\maketag@@@#1{\hbox{\m@th\large\normalfont#1}}
\begin{align}
-\rho_{ij}(t)<e_{ij}(t)<\rho_{ij}(t),~ j=1,\ldots,6~\text{and}~i\in\mathcal{N}\label{eq11}
\end{align}\endgroup
for all $t\geq0$, where $\rho_{ij}(t),~j=1,\ldots,6$ and $i\in\mathcal{N}$ denote the corresponding performance functions. A candidate exponential performance function could be:\vspace{-2mm}
\begingroup\makeatletter\def\f@size{9.5}\check@mathfonts
\def\maketag@@@#1{\hbox{\m@th\large\normalfont#1}}
\begin{equation}
\rho_{ij}(t)=(\rho_{ij,0}-\rho_{ij,\infty})e^{-\lambda t}+\rho_{ij,\infty},~ i\in\mathcal{N}\label{performane:function}
\end{equation}\endgroup
where the constant $\lambda$ dictates the exponential convergence rate, $\rho_{ij,\infty},~i\in\mathcal{N}$ denotes the ultimate bound and $\rho_{ij,0}$ is chosen to satisfy $\rho_{ij,0} > |e_{ij}(0)|,~i\in\mathcal{N}$. Hence, following \cite{10.3389/frobt.2018.00090}, the estimation law is designed as follows:\vspace{-2mm}
\begingroup\makeatletter\def\f@size{9.5}\check@mathfonts
\def\maketag@@@#1{\hbox{\m@th\large\normalfont#1}}
\begin{equation}
\dot{\hat{x}}^{d_i}_{O_j}=k_{ij}\ln\Bigg( \frac{1+ \frac{e_{ij}(t)}{\rho_{ij}(t)}   }{1- \frac{e_{ij}(t)}{\rho_{ij}(t)}   }     \Bigg),k_{ij}>0,~ j=1,\ldots,6\label{eq12}
\end{equation} \endgroup
for $~i\in\mathcal{N}$, from which the followers' estimate $\hat{\bs{x}}^{d_i}_O(t)=[\hat{\bs{x}}^{d_i}_{O_1}(t),\ldots,\hat{\bs{x}}^{d_i}_{O_6}(t)]^\top,~i\in\mathcal{N}$ is calculated via a simple integration. Moreover, differentiating \eqref{eq12} with respect to time, we acquire the desired acceleration signal:\vspace{-2mm}
\begingroup\makeatletter\def\f@size{9.5}\check@mathfonts
\def\maketag@@@#1{\hbox{\m@th\large\normalfont#1}}  
\begin{equation}
\ddot{\hat{x}}^{d_i}_{O_j}=\myfrac[1pt]{2k_{ij}}{1-\Big( \frac{e_{ij}(t)}{\rho_{ij}(t)}  \Big)^2}\myfrac[1pt]{\dot{e}_{ij}(t)\rho_{ij}(t)-e_{ij}(t)\dot{\rho}_{ij}(t)   }{\big(\rho_{ij}(t)\big)^2}\label{eq13}
\end{equation}\endgroup
employing only the velocity $\dot{\bs{x}}_O(t)$ of the object and not its acceleration which is unmeasurable. Based on the aforementioned estimation of the object's desired trajectory profile $\hat{\bs{x}}^{d_i}_O(t)$, $\dot{\hat{\bs{x}}}^{d_i}_O(t)$ and $\ddot{\hat{\bs{x}}}^{d_i}_O(t),~i\in\mathcal{N}$, we can easily derive the corresponding desired trajectory profile for the follower's End-Effector, as follows:\vspace{-2mm}
\begingroup\makeatletter\def\f@size{9.5}\check@mathfonts
\def\maketag@@@#1{\hbox{\m@th\large\normalfont#1}}
\begin{equation}
\arraycolsep=0.4pt\def\arraystretch{1.5}\begin{array}{r@{}l}
&\bs{v}_{O_{F_i}}^{d_i}(t)={\bs{J}^{-1}_O({\bs{\eta}}_{2,O})}  \dot{\hat{\bs{x}}}_O^{d_i}(t)\\
&\dot{\bs{v}}^{d_i}_{O_{F_i}}(t)={\bs{J}^{-1}_O({\bs{\eta}}_{2,O})} \ddot{\hat{\bs{x}}}^{d_i}_O+{\dot{\bs{J}}^{-1}_O({\bs{\eta}}_{2,O})}\dot{\hat{\bs{x}}}^{d_i}_O
\label{eq20}
\end{array}
\end{equation}\endgroup
It is worth noting that the proposed estimator is more robust against trajectory profiles with non-zero acceleration than previous results presented in \cite{Kosuge199717,Kosuge19973373}. In particular, our method guarantees bounded closed loop signals and practical asymptotic stabilization of the estimation errors.  Moreover, the ultimate bounds in \eqref{performane:function} can be set arbitrarily small to a value reflecting the resolution of the measurement devices, thus achieving practical convergence of the estimation errors to zero. Furthermore, the transient response depends on the convergence rate of the performance functions $\rho_{ij}(t),j=1,\ldots,6$ and $i\in \mathcal{N}$ that is directly affected by the parameter $\lambda$.
\section{Simulation Results}
The theoretical findings of this work are verified in a dynamic simulation environment built in MATLAB$^\circledR$ presented in our previous works \cite{Heshmati-alamdari201711197}, with sampling time $0.1~sec$, which is common in a real time operation with an underwater robotic system. The UVMS model considered in the simulations is an AUV equipped with a small 4 DoF manipulator attached at the bow of the vehicle (see Fig.\ref{fig:workspace}). The cooperative transportation is performed by 4 UVMSs grasping the object at its corners. The blue UVMS acts as the leader. Thus, we assume that the desired object's configuration as well as the obstacles' position in the workspace are transferred to the leading UVMS beforehand. The obstacles are modeled as spheres ($1$ m radius) and are located in the workspace in order to complicate the transportation task of the object. In this respect, a Navigation Function is constructed following \eqref{eq8} in order to handle the aforementioned constrained workspace. Since, only the leading UVMS (blue) is aware of the object's desired configuration, the followers will estimate it via the proposed algorithm \eqref{eq12}, by simply observing the motion of the object and without communicating explicitly with the leader. Moreover, the dynamics of the UVMS were affected by external disturbances in the form of slowly time varying sea currents modeled by the corresponding velocities ${v}^c_{x}=0.3\sin(\frac{\pi}{15}t)\frac{m}{s}$ and ${v}^c_{y}=0.3\cos(\frac{\pi}{15}t)\frac{m}{s}$.  Finally, the control gains and the parameters of the proposed estimator were chosen as shown in Table-I and Table-II. \vspace{-4mm}
\begin{figure}[h]
	\centering
	\includegraphics[scale=0.29]{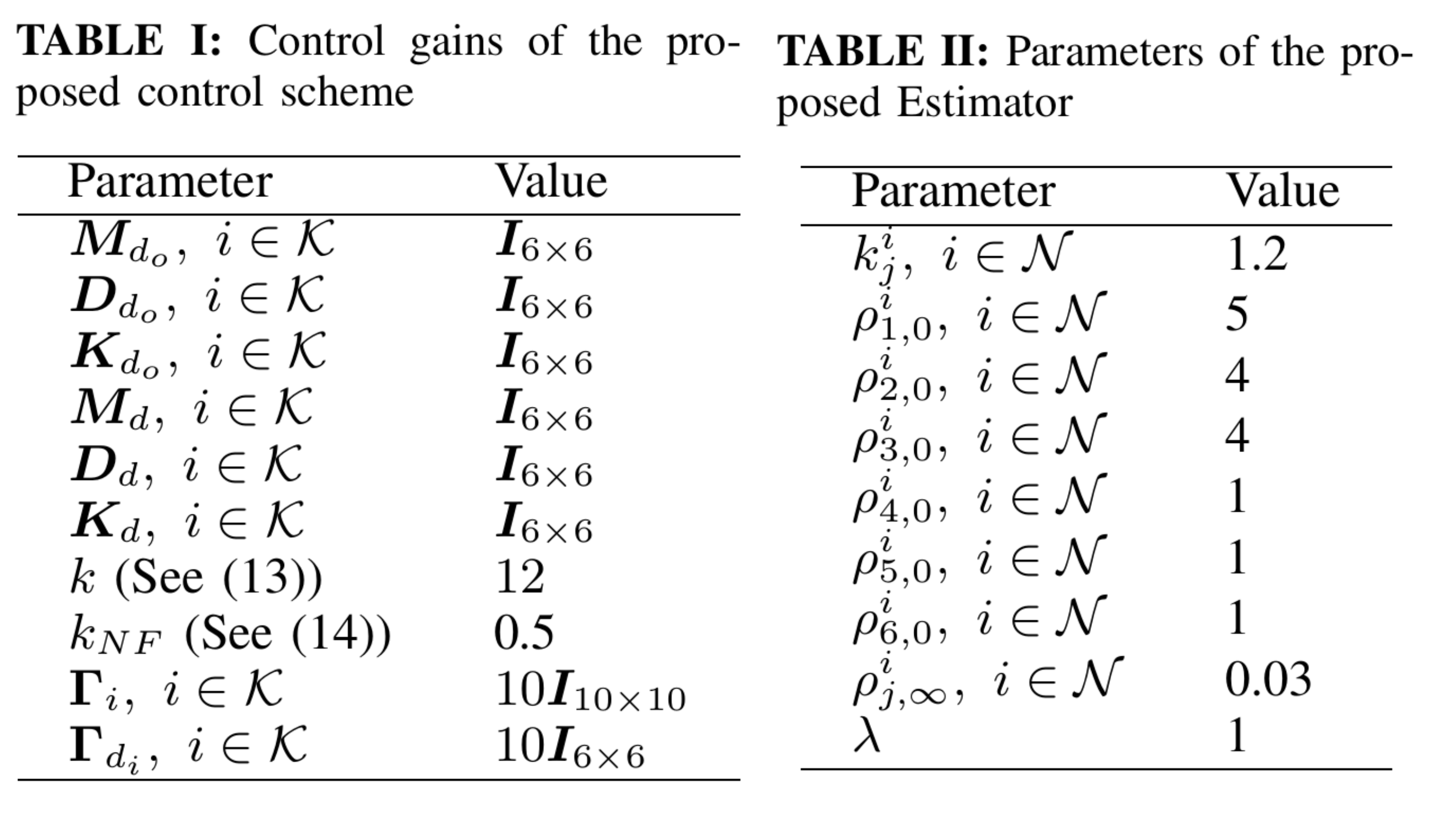}\vspace{-6mm}
\end{figure}
%\begin{table}[h!]
%	\small
%	\centering
%	\setlength\tabcolsep{9pt}
%		\begin{minipage}{0.25\textwidth}
%		\centering
%		\caption{Control gains of the proposed control scheme}
%		\label{controlgains}
%	\begin{tabular}{l l}
%		\hline
%		Parameter &  Value \\
%		\hline
%		$\bs{M}_{d_o},~i\in \mathcal{K}$ & $\bs{I}_{6 \times 6}$ \\
%		$\bs{D}_{d_o},~i\in \mathcal{K}$ & $\bs{I}_{6 \times 6}$ \\
%		$\bs{K}_{d_o},~i\in \mathcal{K}$ & $\bs{I}_{6 \times 6}$\\
%		$\bs{M}_{d},~i\in \mathcal{K}$ & $\bs{I}_{6 \times 6}$ \\
%		$\bs{D}_{d},~i\in \mathcal{K}$ & $\bs{I}_{6 \times 6}$ \\
%		$\bs{K}_{d},~i\in \mathcal{K}$ & $\bs{I}_{6 \times 6}$\\
%		$k$ (See \eqref{eq8})     & $12$\\
%		$k_{NF}$ (See \eqref{eq9}) & 0.5\\
%		$\boldsymbol{\Gamma}_i,~i\in \mathcal{K}$ & $10\bs{I}_{10\times10}$\\ 
%		$\boldsymbol{\Gamma}_{d_i},~i\in \mathcal{K}$ & $10\bs{I}_{6\times6}$\\
%		\hline		
%	\end{tabular}
%	\end{minipage}~
%		\begin{minipage}{0.22\textwidth}
%		\centering
%		\caption{Parameters of the proposed Estimator}
%		\label{pp_param}
%	\begin{tabular}{l l}
%		\hline
%		Parameter &  Value \\
%		\hline
%		$k^i_j,~i\in \mathcal{N}$ & 1.2\\
%		$\rho^i_{1,0},~i\in \mathcal{N}$   & 5 \\
%		$\rho^i_{2,0},~i\in \mathcal{N}$   & 4 \\
%		$\rho^i_{3,0},~i\in \mathcal{N}$   & 4 \\
%		$\rho^i_{4,0},~i\in \mathcal{N}$   & 1 \\
%		$\rho^i_{5,0},~i\in \mathcal{N}$   & 1 \\
%		$\rho^i_{6,0},~i\in \mathcal{N}$   & 1 \\
%		$\rho^i_{j,\infty},~i\in \mathcal{N}$ &   0.03     \\
%		$\lambda$ & 1\\
%		\hline		
%	\end{tabular}
%	\end{minipage}	\vspace{-6mm}
%\end{table}
\subsection*{Simulation Study}
\begin{figure}[t!]
	\centering
	\setlength{\fboxsep}{0pt}%
	\setlength{\fboxrule}{2pt}%
	\includegraphics[scale=0.12]{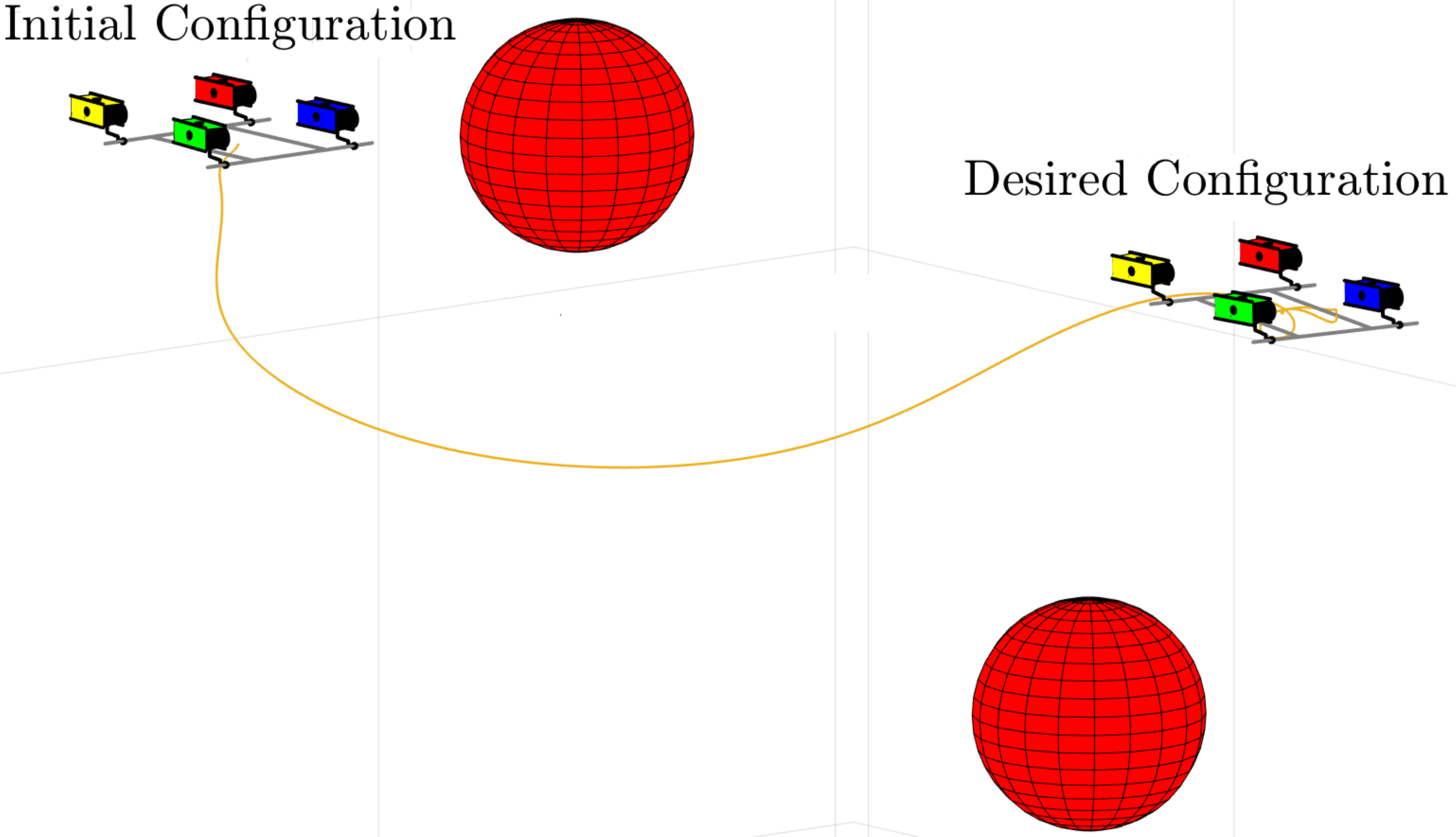}
	\caption{Four UVMSs transport a rigidly grasped object in a constrained workspace with static obstacles. Only the leading UVMS (indicated with blue color) is aware of the object's desired trajectory.}\vspace{-4mm}
	\label{fig:workspace}
\end{figure}

The results are illustrated in Figs.\ref{fig:workspace}-4.  The evolution of the system under the proposed methodology is given in Fig.\ref{fig:workspace}. It should be noticed that the UVMSs have transported cooperatively the grasped object from the initial configuration to the desired one without colliding  with obstacles. By observing the object's tracking error (Fig.3) it can be concluded that even under the influence of external disturbances, the errors in all directions converge very close to zero. 
%\begin{figure}[h]
%	\centering
%	\setlength{\fboxsep}{0pt}%
%	\setlength{\fboxrule}{2pt}%
%	\includegraphics[scale=0.33]{figures/trakingerror1.pdf}
%	\caption{ The object tracking errors in all directions.} \vspace{-4mm}
%	\label{fig:trackingerror1}
%\end{figure}
\begin{figure}[h]
	\centering
	\includegraphics[scale=0.32]{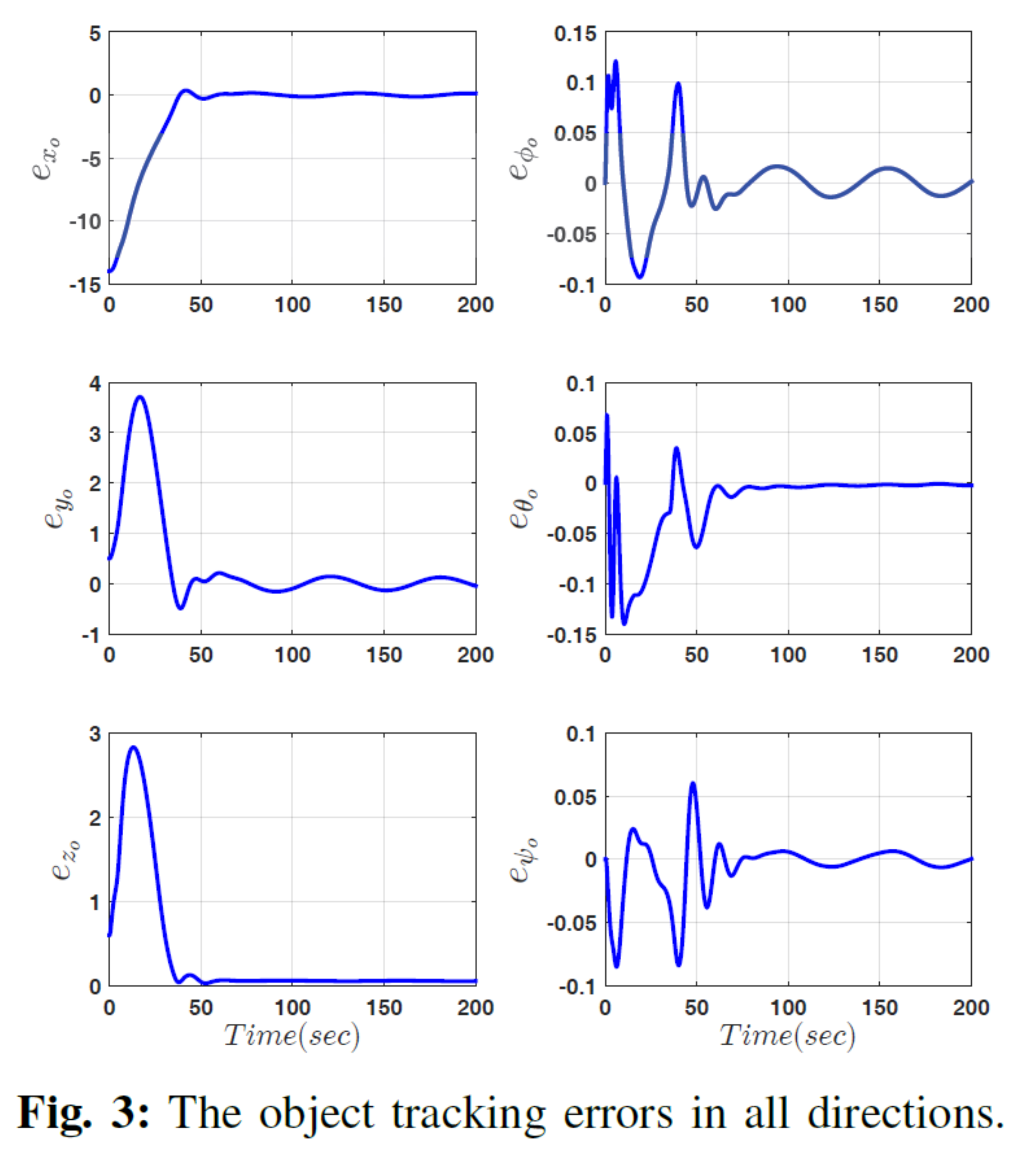}\vspace{-6mm}
\end{figure}
The estimation errors of the proposed estimation scheme are presented in Fig.4. It can be easily seen that the estimation errors  converge smoothly to zero and remain always within the performance envelope defined by the corresponding performance functions as it was expected from the aforementioned theoretical analysis.
%\begin{figure}[h]
%	\centering
%	\setlength{\fboxsep}{0pt}%
%	\setlength{\fboxrule}{2pt}%
%	\includegraphics[scale=0.33]{figures/estimation1.pdf}
%	\caption{ The estimation errors along with the performance bounds imposed by the proposed method. The estimation errors and performance bounds are indicated by blue and red color respectively.}\vspace{-6mm}
%	\label{fig:estimation1}
%\end{figure}
\begin{figure}[h]
	\centering
	\includegraphics[scale=0.36]{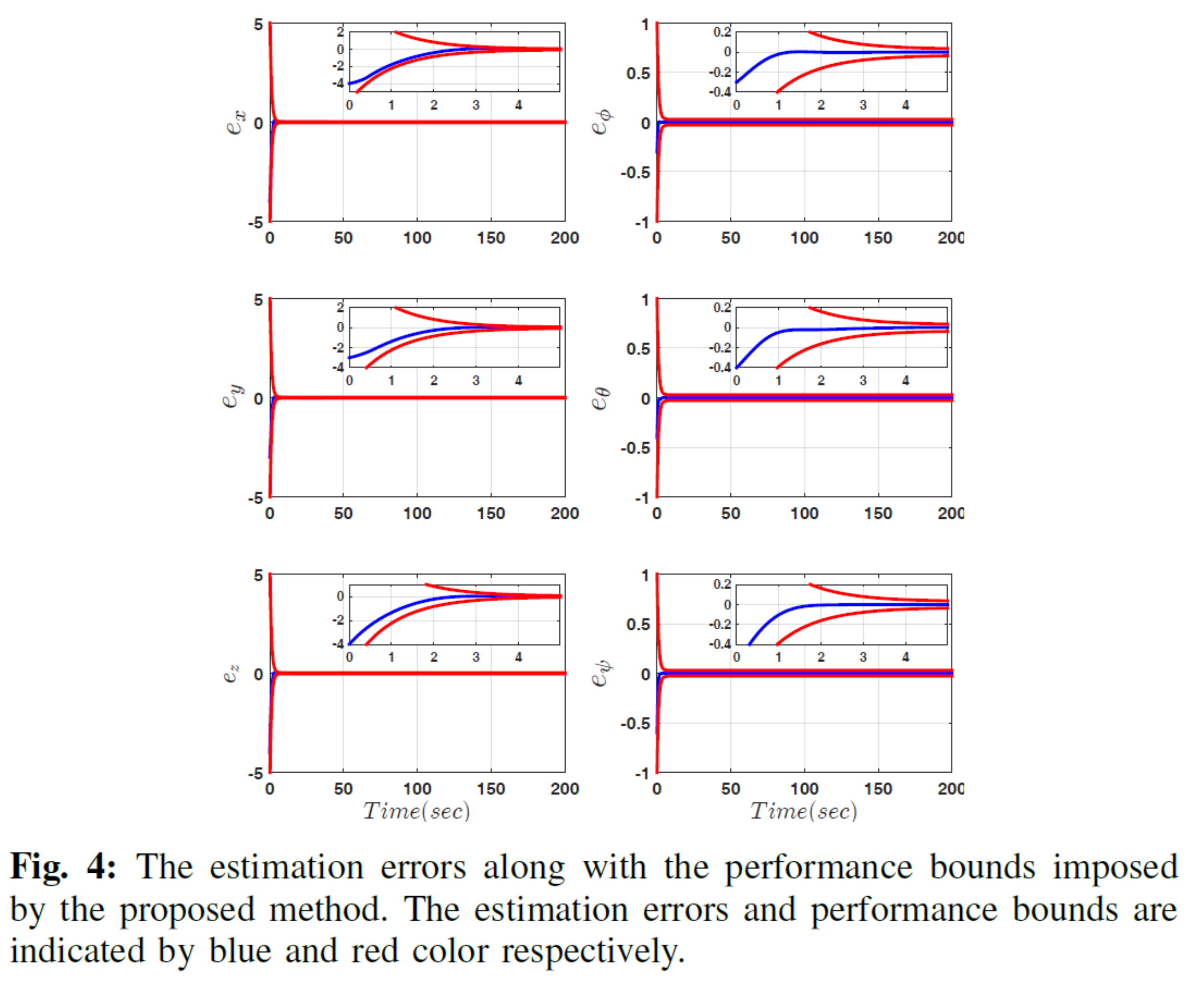}\vspace{-8mm}
\end{figure}
\section{Summary and Future Work}
In this work, we presented a cooperative object transportation scheme for Underwater Vehicle Manipulator Systems under implicit communication, avoiding thus completely tedious explicit data transmission. In the proposed scheme, only the leading UVMS is aware of the desired configuration of the object and  the obstacles' position in the workspace, and aims at navigating safety the team towards the goal configuration. On the contrary, the followers estimate the object's desired trajectory and implement an impedance control law. Moreover, the proposed scheme imposes no restrictions on the underwater communication bandwidth.  Furthermore, the control scheme adopts load sharing among the UVMSs according to their specific payload capabilities. Future research efforts will be devoted towards extending the proposed methodology for multiple UVMSs with underactuated vehicle dynamics.\vspace{-3mm}
{{
		\bibliography{BibCoop}

\begin{thebibliography}{10}

\bibitem{Fossen2}
T.~Fossen, ``Guidance and control of ocean vehicles,'' {\em Wiley, New York,
  1994}.

\bibitem{heshmati2018robust}
S.~Heshmati-alamdari, G.~C. Karras, P.~Marantos, and K.~J. Kyriakopoulos, ``A
  robust model predictive control approach for autonomous underwater vehicles
  operating in a constrained workspace,'' in {\em 2018 IEEE International
  Conference on Robotics and Automation (ICRA)}, pp.~1--5, IEEE, 2018.

\bibitem{heshmati2015robust}
S.~Heshmati-Alamdari, G.~C. Karras, A.~Eqtami, and K.~J. Kyriakopoulos, ``A
  robust self triggered image based visual servoing model predictive control
  scheme for small autonomous robots,'' in {\em Intelligent Robots and Systems
  (IROS), 2015 IEEE/RSJ International Conference on}, pp.~5492--5497, IEEE,
  2015.

\bibitem{HESHMATIALAMDARI2018}
S.~Heshmati-Alamdari, C.~P. Bechlioulis, G.~C. Karras, A.~Nikou, D.~V.
  Dimarogonas, and K.~J. Kyriakopoulos, ``A robust interaction control approach
  for underwater vehicle manipulator systems,'' {\em Annual Reviews in
  Control}, 2018.

\bibitem{Simetti2014364}
E.~Simetti, G.~Casalino, S.~Torelli, A.~Sperindé, and A.~Turetta, ``Floating
  underwater manipulation: Developed control methodology and experimental
  validation within the trident project,'' {\em Journal of Field Robotics},
  vol.~31, no.~3, pp.~364--385, 2014.

\bibitem{Ribas20152583}
D.~Ribas, P.~Ridao, A.~Turetta, C.~Melchiorri, G.~Palli, J.~Fernandez, and
  P.~Sanz, ``I-auv mechatronics integration for the trident fp7 project,'' {\em
  IEEE/ASME Transactions on Mechatronics}, vol.~20, no.~5, pp.~2583--2592,
  2015.

\bibitem{hurtos2014sonar}
N.~Hurtos, N.~Palomeras, A.~Carrera, M.~Carreras, C.~P. Bechlioulis, G.~C.
  Karras, S.~Hesmati-alamdari, and K.~Kyriakopoulos, ``Sonar-based chain
  following using an autonomous underwater vehicle,'' in {\em Intelligent
  Robots and Systems (IROS 2014), 2014 IEEE/RSJ International Conference on},
  pp.~1978--1983, IEEE, 2014.

\bibitem{heshmati2014self}
S.~Heshmati-Alamdari, A.~Eqtami, G.~C. Karras, D.~V. Dimarogonas, and K.~J.
  Kyriakopoulos, ``A self-triggered visual servoing model predictive control
  scheme for under-actuated underwater robotic vehicles,'' in {\em Robotics and
  Automation (ICRA), 2014 IEEE International Conference on}, pp.~3826--3831,
  IEEE, 2014.

\bibitem{Gancet2015218}
J.~Gancet, D.~Urbina, P.~Letier, M.~Ilzokvitz, P.~Weiss, F.~Gauch,
  G.~Antonelli, G.~Indiveri, G.~Casalino, A.~Birk, M.~Pfingsthorn, S.~Calinon,
  A.~Tanwani, A.~Turetta, C.~Walen, and L.~Guilpain, ``Dexrov: Dexterous
  undersea inspection and maintenance in presence of communication latencies,''
  {\em IFAC-PapersOnLine}, vol.~28, no.~2, pp.~218--223, 2015.

\bibitem{Marani200915}
G.~Marani, S.~Choi, and J.~Yuh, ``Underwater autonomous manipulation for
  intervention missions auvs,'' {\em Ocean Engineering}, vol.~36, no.~1,
  pp.~15--23, 2009.

\bibitem{Stilwell20002358}
D.~J. Stilwell and B.~E. Bishop, ``Framework for decentralized control of
  autonomous vehicles,'' {\em In Proceedings of the IEEE International
  Conference on Robotics and Automation}, vol.~3, pp.~2358--2363, 2000.

\bibitem{Nikou2017707}
A.~Nikou, C.~Verginis, S.~Heshmati-Alamdari, and D.~Dimarogonas, ``A nonlinear
  model predictive control scheme for cooperative manipulation with singularity
  and collision avoidance,'' pp.~707--712, 2017.

\bibitem{alex_chris_ECC_2018}
C.~K. Verginis, A.~Nikou, and D.~V. Dimarogonas, ``{C}ommunication-based
  {D}ecentralized {C}ooperative {O}bject {T}ransportation {U}sing {N}onlinear
  {M}odel {P}redictive {C}ontrol,'' {\em European Control Conference (ECC)},
  2018.

\bibitem{Dickson19973589}
W.~C. Dickson, R.~H. Cannon~Jr., and S.~M. Rock, ``Decentralized object
  impedance controller for object/robot-team systems: Theory and experiments,''
  {\em In Proceedings of the IEEE International Conference on Robotics and
  Automation}, vol.~4, pp.~3589--3596, 1997.

\bibitem{Conti2015261}
R.~Conti, E.~Meli, A.~Ridolfi, and B.~Allotta, ``An innovative decentralized
  strategy for i-auvs cooperative manipulation tasks,'' {\em Robotics and
  Autonomous Systems}, vol.~72, pp.~261--276, 2015.

\bibitem{Furferi20161}
R.~Furferi, R.~Conti, E.~Meli, and A.~Ridolfi, ``Optimization of potential
  field method parameters through networks for swarm cooperative manipulation
  tasks,'' {\em International Journal of Advanced Robotic Systems}, vol.~13,
  no.~5, pp.~1--13, 2016.

\bibitem{Simetti2016}
E.~Simetti and G.~Casalino, ``Manipulation and transportation with cooperative
  underwater vehicle manipulator systems,'' {\em IEEE Journal of Oceanic
  Engineering}, 2016.

\bibitem{Manerikar2015}
N.~Manerikar, G.~Casalino, E.~Simetti, S.~Torelli, and A.~Sperinde, ``On
  cooperation between autonomous underwater floating manipulation systems,''
  {\em 2015 IEEE Underwater Technology, UT 2015}, 2015.

\bibitem{Bechlioulis20101220}
C.~Bechlioulis and G.~Rovithakis, ``Prescribed performance adaptive control for
  multi-input multi-output affine in the control nonlinear systems,'' {\em IEEE
  Transactions on Automatic Control}, vol.~55, no.~5, pp.~1220--1226, 2010.

\bibitem{sciavicco2012modelling}
L.~Sciavicco and B.~Siciliano, {\em Modelling and control of robot
  manipulators}.
\newblock Springer Science \& Business Media, 2012.

\bibitem{antonelli}
G.~Antonelli, {\em ``{U}nderwater {R}obots"}.
\newblock Springer Tracts in Advanced Robotics, Springer International
  Publishing, 2013.

\bibitem{Siciliano-b1129198}
B.~Siciliano, L.~Sciavicco, and L.~Villani, {\em Robotics : modelling, planning
  and control}.
\newblock Advanced Textbooks in Control and Signal Processing, Springer, 2009.
\newblock 013-81159.

\bibitem{Slotine1987595}
J.-J.~E. Slotine and W.~Li, ``Adaptive strategies in constrained
  manipulation.,'' {\em In Proceedings of the IEEE International Conference on
  Robotics and Automation.}, pp.~595--601, 1987.

\bibitem{Tatlicioglu2009873}
E.~Tatlicioglu, D.~Braganza, T.~Burg, and D.~Dawson, ``Adaptive control of
  redundant robot manipulators with sub-task objectives,'' {\em Robotica},
  vol.~27, no.~6, pp.~873--881, 2009.

\bibitem{Koditschek1990412}
D.~Koditschek and E.~Rimon, ``Robot navigation functions on manifolds with
  boundary,'' {\em Advances in Applied Mathematics}, vol.~11, no.~4,
  pp.~412--442, 1990.

\bibitem{de2005sensorless}
A.~De~Luca and R.~Mattone, ``Sensorless robot collision detection and hybrid
  force/motion control,'' in {\em Robotics and Automation, 2005. ICRA 2005.
  Proceedings of the 2005 IEEE International Conference on}, pp.~999--1004,
  IEEE, 2005.

\bibitem{caccavale1998resolved}
F.~Caccavale, C.~Natale, B.~Siciliano, and L.~Villani, ``Resolved-acceleration
  control of robot manipulators: A critical review with experiments,'' {\em
  Robotica}, vol.~16, no.~5, pp.~565--573, 1998.

\bibitem{10.3389/frobt.2018.00090}
C.~P. Bechlioulis and K.~J. Kyriakopoulos, ``Collaborative multi-robot
  transportation in obstacle-cluttered environments via implicit
  communication,'' {\em Frontiers in Robotics and AI}, vol.~5, p.~90, 2018.

\bibitem{Kosuge199717}
K.~Kosuge, T.~Oosumi, and H.~Seki, ``Decentralized control of multiple
  manipulators handling an object in coordination based on impedance control of
  each arm,'' {\em IEEE International Conference on Intelligent Robots and
  Systems}, vol.~1, pp.~17--22, 1997.

\bibitem{Kosuge19973373}
K.~Kosuge, T.~Oosumi, and K.~Chiba, ``Load sharing of decentralized-controlled
  multiple mobile robots handling a single object,'' {\em In Proceedings of the
  IEEE International Conference on Robotics and Automation}, vol.~4, 1997.

\bibitem{Heshmati-alamdari201711197}
S.~Heshmati-alamdari, A.~Nikou, K.~Kyriakopoulos, and D.~Dimarogonas, ``A
  robust force control approach for underwater vehicle manipulator systems,''
  {\em IFAC-PapersOnLine}, vol.~50, no.~1, pp.~11197--11202, 2017.

\end{thebibliography}
		\bibliographystyle{ieeetr}
}}
\end{document}